\pdfoutput=1

\documentclass[11pt]{article}

\usepackage[]{acl}

\usepackage{times}
\usepackage{latexsym}
\usepackage{url}
\usepackage{amssymb}
\usepackage[T1]{fontenc}

\usepackage[utf8]{inputenc}

\usepackage{microtype}
\usepackage{booktabs}
\usepackage{multirow}

\usepackage{inconsolata}

\usepackage{graphicx}
\usepackage{multirow}

\title{Who Taught You That? \\ Tracing Teachers in Model Distillation}

\author{\textbf{Somin Wadhwa}$^{*}$\qquad \textbf{Chantal Shaib}$^{*}$\qquad \textbf{Silvio Amir}\qquad \textbf{Byron C. Wallace}\qquad \\ 
Northeastern University \\ 
\texttt{\{wadhwa.s, shaib.c, s.amir, b.wallace\}@northeastern.edu} 
}

\begin{document}
\maketitle
\begingroup\def\thefootnote{*}\footnotetext{Equal Contribution.}\endgroup
\begin{abstract}
Model distillation---using outputs from a large teacher model to teach a small student model---is a practical means of creating efficient models for a particular task. 
We ask: Can we identify a students' teacher based on its outputs? 
Such ``footprints'' left by teacher LLMs would be interesting artifacts. 
Beyond this, reliable teacher inference may have practical implications as actors seek to distill specific capabilities of massive proprietary LLMs into deployed smaller LMs, potentially violating terms of service.\footnote{Consider, e.g., speculation as to if DeepSeek was at least partly distilled from OpenAI's ChatGPT \cite{apnews2025deepseek}.}   
We consider practical task distillation targets including summarization, question answering, and instruction-following. 
We assume a finite set of candidate teacher models, which we treat as blackboxes. 
We design discriminative models that operate over lexical features. 
We find that $n$-gram similarity alone is unreliable for identifying teachers, but \emph{part-of-speech (PoS)} templates \cite{shaib-etal-2024-detection} preferred by student models mimic those of their teachers. 
\end{abstract}

\begin{figure}
\centering
   \includegraphics[scale=0.40]{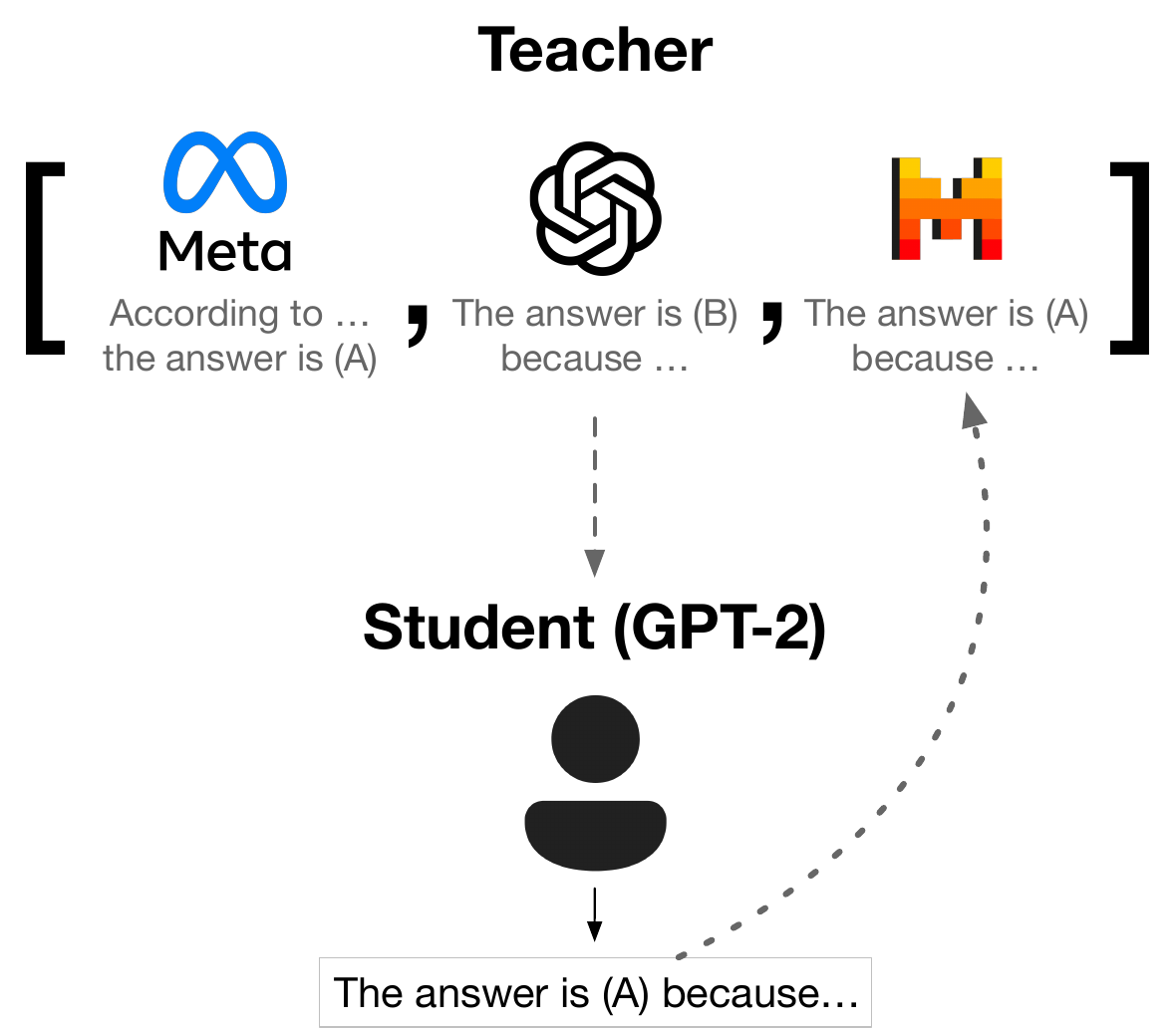}
  \caption{We introduce the problem of \emph{teacher model attribution}: Given a distilled student model (e.g., a fine-tuned GPT-2), determine which of a set of possible teacher models was distilled (here, Mistral).}
  \label{fig:the_problem}
\end{figure}

\section{Introduction}

\emph{Model distillation} \cite{buciluǎ2006model,hinton2015distilling} entails teaching a small model using outputs sampled from a larger model. 
In the LLM era, this has proven an especially practical strategy: Distillation can imbue efficient small language models with task-specific capabilities competitive with (expensive) teacher LLMs \cite{xu2024survey}. 
In this work we assess the degree to which teachers inculcate ``signatures'' into student models during distillation.  
Specifically, we ask: \textbf{Given a \textit{distilled student model} can we 
identify its \emph{teacher} from a candidate set} (Figure \ref{fig:the_problem})?

This may have practical implications. 
Imagine a start-up company distilling a particular piece of functionality (summarization, say) from a large proprietary model and using this to power a paid service. 
This may violate terms of service, so LLM providers might be keen to identify such cases. 


We consider a set of tasks for which distillation has proven successful in prior work. 
For reasoning and math tasks, distillation targets often include ``reasoning'' elicited from the teacher, as this has been shown to improve student performance substantially \cite{ho-etal-2023-large,li-etal-2023-symbolic,shridhar-etal-2023-distilling,wadhwa-etal-2024-investigating}.
We also consider broader ``distillation'', namely general instruction-tuning using examples elicited from a massive LLM, as done by Alpaca \cite{alpaca}. 

One might think that simply measuring similarity between the outputs generated by a student model and candidate teachers
would suffice to identify the teacher. 
However, we find that basic similarity measurements over texts provide practically no useful signal in this respect. 
Another obvious strategy would be to evaluate the likelihood of student outputs under different teacher models. 
Intuitively, the teacher should prefer (assign high likelihood to) outputs generated by its pupil. 
But again we find that such teacher perplexities are insufficient to reliably pick a teacher from a candidate set.

If shallow measures of similarity do not suffice to identify teachers, what might? 
We explore syntactic Part-of-Speech (PoS) templates \cite{shaib-etal-2024-detection} as an alternative means of identifying teachers via higher-order lexical features.
We find that these templates---which are more abstract than raw output texts---carry relatively strong signal about the teacher used in distillation.

\section{Problem Setup}
\label{section:setup}




Distilling LLMs involves training a small, cost-efficient student model (e.g., GPT-2) using outputs from a much larger teacher model (e.g., GPT-4) to create an efficient yet capable model. We assume a student model $m$ trained on data from one of several possible teacher LLMs, \( {\mathcal{M} = \{ M_1, M_2, \dots, M_T \} }\).\footnote{In practice, a limited number of proprietary LLMs (e.g., GPT-*, Claude, or Llama) are typically used for distillation.} Our goal is to identify which teacher model $M_j$ trained $m$, without access to the original distillation data.

In our experiments, we use fine-tuned {\tt GPT-2} \cite{radford2019language} and Olmo-1B \cite{Groeneveld2023OLMo} as student models ($m$) and a teacher set $\mathcal{M}$ = \{{\tt Llama3-8B}, {\tt Llama3-70B}, {\tt Mistral-7B}, {\tt Mixtral}, {\tt Gemma2-9B}\}, selecting open models for reproducibility. The teachers range from 8B to 70B parameters, all significantly larger than $m$.


\begin{figure*}
    \centering
    \includegraphics[scale=0.63]{
      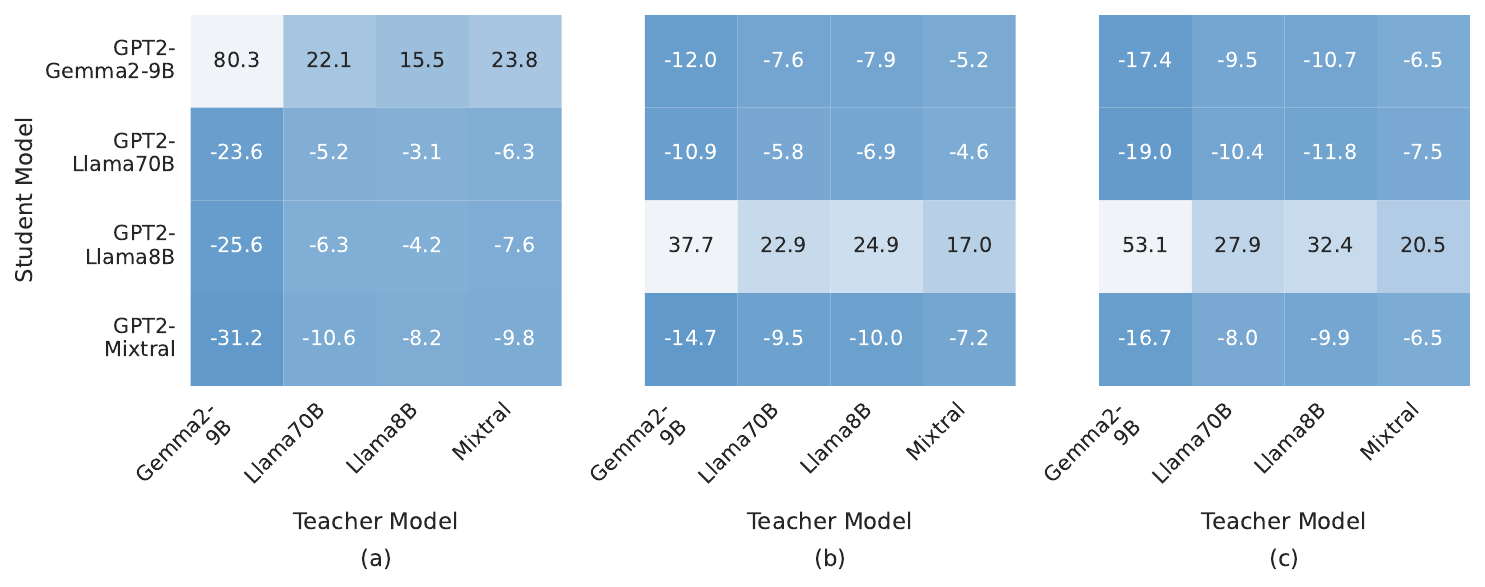}
    \caption{Perplexity under teacher models of texts generated by different pupils on (a) Rotten-Tomatoes, (b) QuaRel, and (c) OpenBookQA. Teacher perplexity does not consistently identify the teacher.}
    \label{fig:ppl_fig}
\end{figure*}

\paragraph{Tasks and Datasets} 
We experiment with tasks where larger models have been successfully distilled, including summarization, question answering, and general instruction following. 

For summarization, we use CNN-DailyMail \cite{see-etal-2017-get}, Rotten Tomatoes \cite{leone2020rotten}, and PubMed \cite{bharti2020}. For question answering, we use OpenbookQA \cite{mihaylov-etal-2018-suit} and CommonsenseQA \cite{talmor-etal-2019-commonsenseqa}. For instruction following, we randomly sample 10K instances from Alpaca~\cite{alpaca}.\footnote{Downsampling details in Appendix \ref{appx:data}.}

Training data is generated from all five teacher models. For summarization and instruction following, teachers generate outputs directly. For question answering, following \citet{li-etal-2023-symbolic}, we generate \textit{reasoning} chains for correct labels, using both chains and labels as student targets.

\section{Teacher Attribution Methods}

Here we describe models we considered to identify a teacher model based on a set of student outputs.
These include approaches based on: (i) perplexity; (ii) similarity metrics; and (iii) syntactic patterns.

LLMs ``prefer'' (assign lower perplexities to) text that they have produced. 
This suggests that simply comparing perplexities of student-generated text under each candidate teacher may be a viable identification strategy: 
We would expect the true teacher to assign lower perplexity to outputs from their pupils. 
However, as shown in Figure \ref{fig:ppl_fig}, this is insufficient to discriminate between teacher models. 
For example, Gemma assigns much higher perplexities to summaries produced by a model distilled from its own outputs (\ref{fig:ppl_fig}, a). 

\subsection{Similarity Metrics}

\begin{table}[t]
\renewcommand*{\arraystretch}{1.5}
\small
\centering
\begin{tabular}{p{1.5cm}p{2cm}cc}
\hline
\multicolumn{1}{c}{}                                                         & \multicolumn{1}{l}{\textbf{Teacher}} & \textbf{BoW} & \textbf{BERTScore} \\ \hline
\multirow{6}{*}{\texttt{\textbf{}}OBQA}                                     & \textcolor{blue}{\textbf{Llama8B}}
& 0.54 & 0.65 \\
                                                                            & Llama-70B  \textcolor{red}{$\times$}            & \textbf{0.56} & \textbf{0.71} \\
                                                                            &  Mistral-7B             & 0.53 & 0.62 \\
                                                                            &   Mixtral \textcolor{red}{$\times$}             & \textbf{0.56} & 0.65 \\
                                                                            &   Gemma2-9B             & 0.51 & 0.49 \\ 
                                                                            \hline
\multirow{6}{*}{\texttt{\textbf{}}Alpaca}                               &  \textcolor{blue}{\textbf{Llama8B}}
& 0.26 & 0.21 \\
                                                                            &   Llama-70B             & 0.22 & 0.25 \\
                                                                            &   Mistral-7B \textcolor{red}{$\times$}           & 0.25 & \textbf{0.26} \\
                                                                            &   Mixtral  \textcolor{red}{$\times$}             & \textbf{0.27} & \textbf{0.26} \\
                                                                            &  Gemma2-9B              & 0.19 & 0.11 \\   \hline
\multirow{6}{*}{\texttt{\textbf{}}C-D}                        & \textcolor{blue}{\textbf{Llama8B~\checkmark}}                      & \textbf{0.71} & 0.67 \\
                                                                            &   Llama-70B \textcolor{red}{$\times$}            & 0.60 & \textbf{0.68} \\
                                                                            & Mistral-7B              & 0.43 & 0.49 \\
                                                                            &  Mixtral                & 0.41 & 0.48 \\
                                                                            &  Gemma2-9B              & 0.28 & 0.31 \\   \hline
\end{tabular}
\caption{Neither cosine similarities (BoW) nor BERTScores between student (fine-tuned GPT-2) and candidate teachers reliably reveal the true teacher (Llama8B).}
\label{tab:results_similarity}
\end{table}

One approach to matching students with their teachers is to measure the similarity between the texts that they generate. 
We would expect texts generated by a student model to resemble those produced by its teacher (as compared to other LLMs).

We use BERTscore \cite{zhang2020bertscoreevaluatingtextgeneration} and cosine similarity based on a bag-of-words representations as similarity measures. 
Table \ref{tab:results_similarity}\footnote{Additional results in Appendix Figure \ref{fig:sim_fig}.} shows the similarity between student and teacher outputs, as compared with other candidate teacher models across each dataset. 
Raw similarities are unreliable indicators of teacher models.

\begin{figure}[t]
\centering
  \includegraphics[scale=0.44]{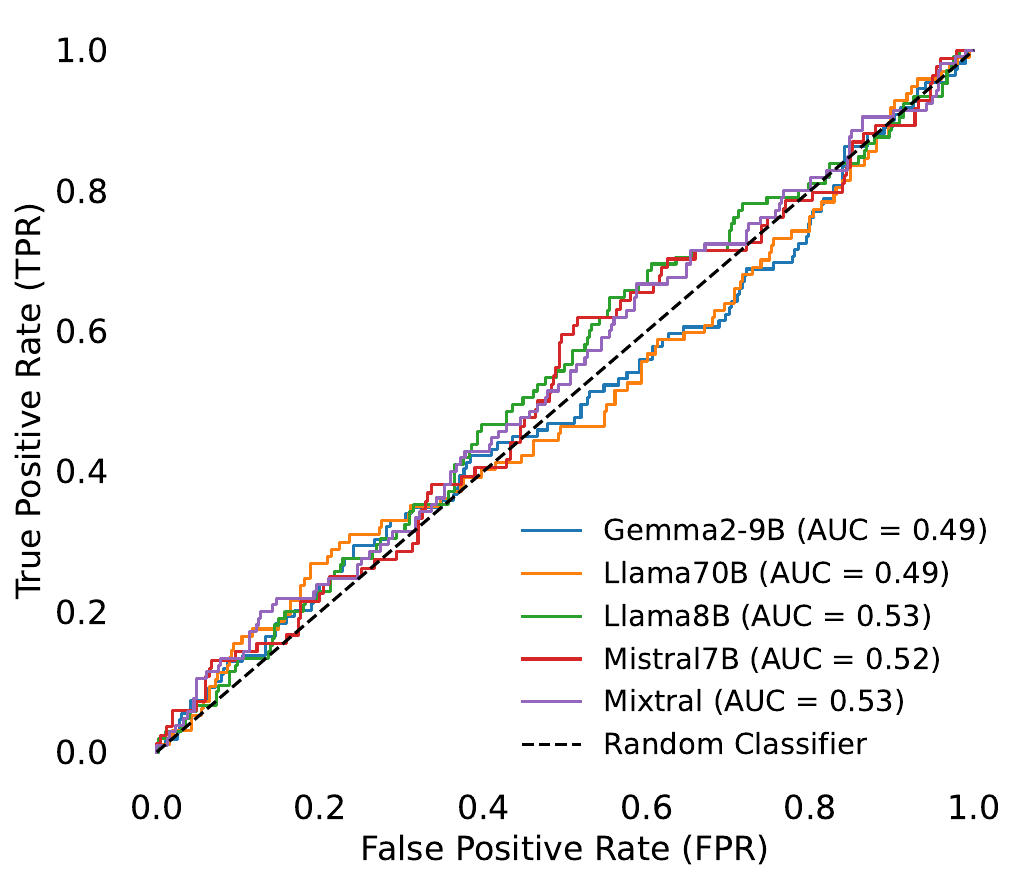}
  \caption{AUC-ROC curves for a one-vs-rest LR classifier using similarity score as the sole feature. Performance across models is close to random (AUC $\approx$ 0.49–0.53), indicating limited discriminative power.}
  \label{fig:sim_auc}
\end{figure}

This limitation is further highlighted in Figure \ref{fig:sim_auc}, which presents AUC-ROC curves for logistic regression models using similarity scores as standalone features in a one-vs-rest classification setup. The average AUC score hovers around 0.52, indicating that similarity scores alone provide little discriminatory power in distinguishing between teacher models. These findings emphasize the need for more robust approaches to accurately differentiate teacher-student relationships.

\begin{figure*}
    \centering
    \includegraphics[scale=0.21]{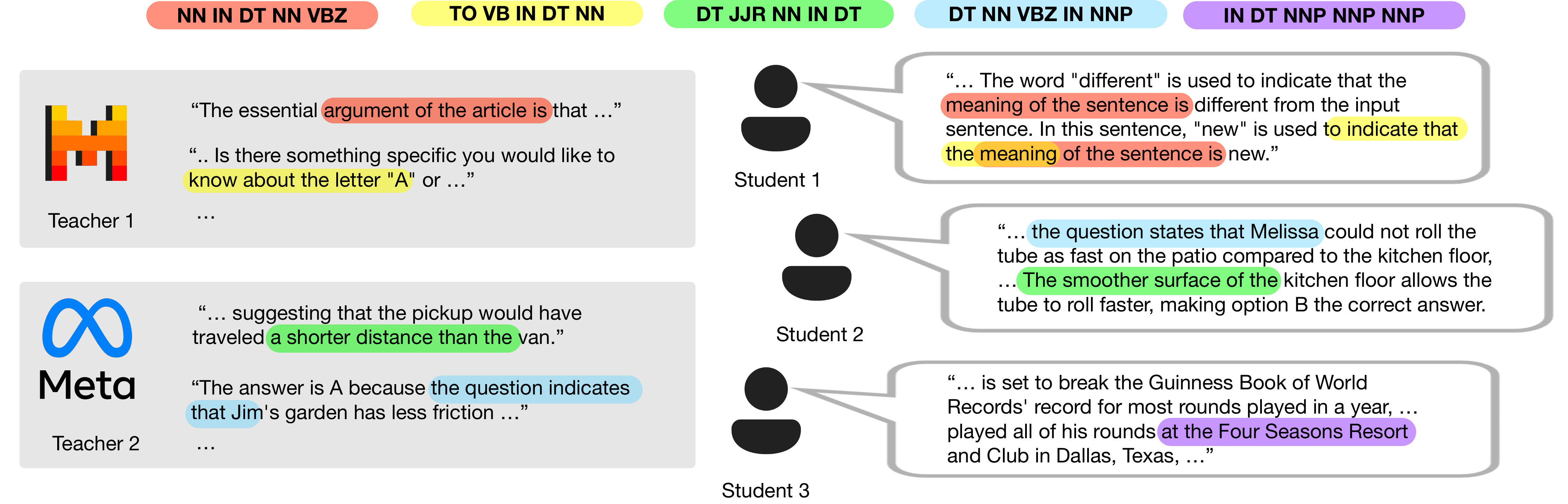}
    \caption{Influence of teacher models on student outputs, highlighting the retention of Part-of-Speech (PoS) templates. The color-coded PoS sequences illustrate how students inherit structural patterns from their respective teachers, suggesting that syntactic characteristics are preserved to some extent during knowledge transfer. This pattern indicates that PoS templates can serve as a distinguishing feature in identifying which teacher model was used to train a given student.}
    \label{fig:pos_label}
\end{figure*}

\begin{table*}[t]
\normalsize
\centering
\addtolength{\tabcolsep}{5pt} 
\begin{tabular}{@{}llccccccc@{}}
\toprule

                                &     & \multicolumn{1}{l}{C-D} & \multicolumn{1}{l}{P-M} & \multicolumn{1}{l}{R-T} & \multicolumn{1}{l}{CSQA} & \multicolumn{1}{l}{OBQA} & \multicolumn{1}{l}{QRe} & \multicolumn{1}{l}{Alpaca} \\ \midrule

\multirow{3}{*}{GPT-2}   & BERT          & 0.46            & 0.55       & 0.40        & 0.44     & 0.38         & 0.35                    & 0.51                       \\
& $n$-grams (1-4)  & 0.58         & 0.68       & 0.44        & 0.56     & 0.48         & 0.50            & \textbf{0.56 }                      \\
& PoS Templates & \textbf{0.60} & \textbf{0.71} & \textbf{0.54} & \textbf{0.69 } & \textbf{0.51} & \textbf{0.59 }  & 0.55                       \\  \midrule
\multirow{3}{*}{Olmo-1B} & BERT          & 0.45                    & 0.65                    & 0.41                    & 0.40                     & 0.42                     & 0.31                    & 0.46                       \\
                         & n-grams       & 0.60                    & 0.62                    & \textbf{0.48}                    & 0.55                     & 0.42                     & 0.58                    & 0.50                       \\
                         & POS Templates & \textbf{0.61}                    & \textbf{0.74}                    & 0.45                    & \textbf{0.59}                     & \textbf{0.43}                     & \textbf{0.61}                    & \textbf{0.53}                       \\ \bottomrule

\end{tabular}
\addtolength{\tabcolsep}{5pt} 
\caption{Classification performance of logistic regression and BERT using different feature representations across datasets. Models perform best when using PoS template features. 
This trend is consistent across different datasets, suggesting that syntactic structures (captured by PoS Templates) and higher-order lexical patterns (captured by $n$-grams) provide more discriminative power compared to simple word occurrence (BoW).}
\label{tab:main_res}
\end{table*}

\subsection{Syntactic Patterns}

We next test whether ``syntactic templates'' (PoS sequences) provide signal sufficient to distinguish between different student models, inspired by recent work showing that LLMs prefer certain sentence constructions, more so than human authors \citep{shaib-etal-2024-detection}. 
Student models may internalize such structures, providing a signature of their teacher. 


We extract PoS templates with the {\tt diversity} package, \cite{shaib2024standardizing}\footnote{\url{https://pypi.org/project/diversity/}} finding the 50 most common PoS patterns of length 4 across all teachers, for up to 200 test instances per model. 
This yields a set of unique PoS patterns and corresponding sentences. 
We construct a training dataset using PoS template indicators as features and teacher models as targets. 
We train a simple logistic regression classifier and evaluate its performance on test sets generated by student models.

Results are summarized in Table \ref{tab:main_res}. PoS templates consistently outperform $n$-gram (up to n=4) 
and BERT-based models in distinguishing between teachers. 
For instance, when using GPT-2 as the student model, on  PubMed the PoS template-based classifier achieves 0.68 accuracy, compared to 0.61 with $n$-grams. 
Similarly, on CommonsenseQA data PoS templates yield 0.67 accuracy, vs. 0.48 for $n$-grams. However, on Alpaca data $n$-grams slightly outperform PoS templates (0.52 vs. 0.48), demonstrating a marginal exception to the trend. Overall, these findings highlight the robustness of syntactic patterns as features for teacher model classification, particularly when compared to simpler word-based representations.

\section{Related Work}

\paragraph{Distillation with LLMs}

Smaller models can learn from explanations (i.e., rationales) \citep{hase-bansal-2022-models} despite lacking inherent \textit{step-by-step} reasoning \cite{10.5555/3600270.3602070}. However, they can be trained to generate such chains \cite{magister-etal-2023-teaching}. Recent studies \citep{10.1007/978-3-031-77844-5_13,ho-etal-2023-large} show that rationales as training signals significantly improve student models. \citet{li-etal-2023-symbolic} explore factors influencing \textit{teacher} corpus creation for commonsense reasoning, while \citet{pmlr-v202-fu23d} examine trade-offs between generalizability and CoT-generation, highlighting rationale quality as crucial for performance. More recently, \citet{wadhwa-etal-2024-investigating} found that placing rationales \emph{after} labels enhances CoT-augmented distillation. 

\paragraph{Origin Tracing} 

Data provenance techniques like watermarking can trace the origins of distilled models. \citet{li2024statistical} introduce statistical tests for identifying source models without requiring access to model probabilities. \citet{li2024identifying} adopt generation-time watermarking, though this assumes student training data is watermarked. 
Most relevant to our work, \citet{li2023origin} use perplexity and contrastive training for text origin detection, whereas we investigate linguistic features as model signatures.

\section{Conclusions}
We have introduced the problem of identifying the teacher model used to train a distilled student model. 
We demonstrated that standard approaches, such as measuring similarity between teacher and student outputs or using perplexity as a proxy, are insufficient for reliable attribution. 
We introduced syntactic part-of-speech (PoS) templates as higher-order linguistic features capable of capturing distinctive signals from teacher models that persist in distilled student outputs.

We find that these linguistic patterns provide comparatively strong signal about teachers across tasks and datasets. 
This does not assume access to teacher model internals or use of watermarking strategies. 
While this approach achieves accuracy well above chance (which would be 0.2 here), it leaves considerable room for improvement, and the practical implications of the accuracies we have achieved are a bit unclear. 

Nonetheless, this work lays the foundation for further studies in teacher attribution, with potential implications for understanding model behaviors, ensuring compliance with usage agreements, and enhancing the transparency of AI systems. 
Future work could explore extensions to other types of linguistic features, investigate more complex attribution scenarios, or develop methods to counteract attribution in privacy-sensitive applications.

\section*{Limitations}
This work has several important limitations.

First, the effectiveness of syntactic templates as distinguishing features is dependent on the extent to which a student model retains the linguistic patterns of its teacher. 
This retention may be affected by factors such as additional fine-tuning, data augmentation, multi-teacher distillation, or \textit{shared} footprints among different teachers that are trained on the same data, which could obscure attribution signals. Investigating how these factors impact our attribution framework is an important direction for future research.

Second, while we show that syntactic patterns provide stronger attribution signals than traditional similarity metrics, our results indicate that there is considerable room for improvement.
While the predictive accuracies we reported are well above chance, they are far from perfect. 
Future work could explore integrating multiple complementary signals, such as semantic embeddings, paraphrase consistency, or latent-space representations, to further improve attribution accuracy.

Finally, our approach assumes a closed-set identification scenario where the true teacher model is among a predefined set of candidate models. 
Extending our methods to accommodate an arbitrarily large set of candidate teacher models is therefore an open direction for future work. 

\bibliography{custom}

\begin{thebibliography}{29}
\providecommand{\natexlab}[1]{#1}

\bibitem[{{Associated Press}(2025)}]{apnews2025deepseek}
{Associated Press}. 2025.
\newblock \href {https://apnews.com/article/deepseek-ai-chatgpt-openai-copyright-a94168f3b8caa51623ce1b75b5ffcc51} {Did deepseek copy chatgpt to make new ai chatbot?}
\newblock \emph{AP News}.
\newblock Accessed: 2025-01-30.

\bibitem[{Buciluǎ et~al.(2006)Buciluǎ, Caruana, and Niculescu-Mizil}]{buciluǎ2006model}
Cristian Buciluǎ, Rich Caruana, and Alexandru Niculescu-Mizil. 2006.
\newblock Model compression.
\newblock In \emph{Proceedings of the 12th ACM SIGKDD international conference on Knowledge discovery and data mining}, pages 535--541.

\bibitem[{Fu et~al.(2023)Fu, Peng, Ou, Sabharwal, and Khot}]{pmlr-v202-fu23d}
Yao Fu, Hao Peng, Litu Ou, Ashish Sabharwal, and Tushar Khot. 2023.
\newblock \href {https://proceedings.mlr.press/v202/fu23d.html} {Specializing smaller language models towards multi-step reasoning}.
\newblock In \emph{Proceedings of the 40th International Conference on Machine Learning}, volume 202 of \emph{Proceedings of Machine Learning Research}, pages 10421--10430. PMLR.

\bibitem[{Groeneveld et~al.(2024)Groeneveld, Beltagy, Walsh, Bhagia, Kinney, Tafjord, Jha, Ivison, Magnusson, Wang, Arora, Atkinson, Authur, Chandu, Cohan, Dumas, Elazar, Gu, Hessel, Khot, Merrill, Morrison, Muennighoff, Naik, Nam, Peters, Pyatkin, Ravichander, Schwenk, Shah, Smith, Subramani, Wortsman, Dasigi, Lambert, Richardson, Dodge, Lo, Soldaini, Smith, and Hajishirzi}]{Groeneveld2023OLMo}
Dirk Groeneveld, Iz~Beltagy, Pete Walsh, Akshita Bhagia, Rodney Kinney, Oyvind Tafjord, Ananya~Harsh Jha, Hamish Ivison, Ian Magnusson, Yizhong Wang, Shane Arora, David Atkinson, Russell Authur, Khyathi Chandu, Arman Cohan, Jennifer Dumas, Yanai Elazar, Yuling Gu, Jack Hessel, Tushar Khot, William Merrill, Jacob Morrison, Niklas Muennighoff, Aakanksha Naik, Crystal Nam, Matthew~E. Peters, Valentina Pyatkin, Abhilasha Ravichander, Dustin Schwenk, Saurabh Shah, Will Smith, Nishant Subramani, Mitchell Wortsman, Pradeep Dasigi, Nathan Lambert, Kyle Richardson, Jesse Dodge, Kyle Lo, Luca Soldaini, Noah~A. Smith, and Hannaneh Hajishirzi. 2024.
\newblock Olmo: Accelerating the science of language models.
\newblock \emph{Preprint}.

\bibitem[{Gupta et~al.(2021)Gupta, Bharti, Nokhiz, and Karnick}]{bharti2020}
Vivek Gupta, Prerna Bharti, Pegah Nokhiz, and Harish Karnick. 2021.
\newblock \href {https://vgupta123.github.io/docs/121_paper.pdf} {Sumpubmed: Summarization dataset of pubmed scientific article}.
\newblock In \emph{Proceedings of the 2021 Conference of the Association for Computational Linguistics: Student Research Workshop}. Association for Computational Linguistics.

\bibitem[{Hase and Bansal(2022)}]{hase-bansal-2022-models}
Peter Hase and Mohit Bansal. 2022.
\newblock \href {https://doi.org/10.18653/v1/2022.lnls-1.4} {When can models learn from explanations? a formal framework for understanding the roles of explanation data}.
\newblock In \emph{Proceedings of the First Workshop on Learning with Natural Language Supervision}, pages 29--39, Dublin, Ireland. Association for Computational Linguistics.

\bibitem[{Hinton(2015)}]{hinton2015distilling}
Geoffrey Hinton. 2015.
\newblock Distilling the knowledge in a neural network.
\newblock \emph{arXiv preprint arXiv:1503.02531}.

\bibitem[{Ho et~al.(2023)Ho, Schmid, and Yun}]{ho-etal-2023-large}
Namgyu Ho, Laura Schmid, and Se-Young Yun. 2023.
\newblock \href {https://doi.org/10.18653/v1/2023.acl-long.830} {Large language models are reasoning teachers}.
\newblock In \emph{Proceedings of the 61st Annual Meeting of the Association for Computational Linguistics (Volume 1: Long Papers)}, pages 14852--14882, Toronto, Canada. Association for Computational Linguistics.

\bibitem[{Leone(2020)}]{leone2020rotten}
Stefano Leone. 2020.
\newblock Rotten tomatoes movies and critic reviews dataset.

\bibitem[{Li et~al.(2023{\natexlab{a}})Li, Wang, Ren, Sun, and Qiu}]{li2023origin}
Linyang Li, Pengyu Wang, Ke~Ren, Tianxiang Sun, and Xipeng Qiu. 2023{\natexlab{a}}.
\newblock Origin tracing and detecting of llms.
\newblock \emph{arXiv preprint arXiv:2304.14072}.

\bibitem[{Li et~al.(2023{\natexlab{b}})Li, Hessel, Yu, Ren, Chang, and Choi}]{li-etal-2023-symbolic}
Liunian~Harold Li, Jack Hessel, Youngjae Yu, Xiang Ren, Kai-Wei Chang, and Yejin Choi. 2023{\natexlab{b}}.
\newblock \href {https://doi.org/10.18653/v1/2023.acl-long.150} {Symbolic chain-of-thought distillation: Small models can also {``}think{''} step-by-step}.
\newblock In \emph{Proceedings of the 61st Annual Meeting of the Association for Computational Linguistics (Volume 1: Long Papers)}, pages 2665--2679, Toronto, Canada. Association for Computational Linguistics.

\bibitem[{Li et~al.(2024{\natexlab{a}})Li, Bai, and Cheng}]{li2024identifying}
Liying Li, Yihan Bai, and Minhao Cheng. 2024{\natexlab{a}}.
\newblock Where am i from? identifying origin of llm-generated content.
\newblock In \emph{Proceedings of the 2024 Conference on Empirical Methods in Natural Language Processing}, pages 12218--12229.

\bibitem[{Li et~al.(2024{\natexlab{b}})Li, Ruan, Wang, Long, and Su}]{li2024statistical}
Xiang Li, Feng Ruan, Huiyuan Wang, Qi~Long, and Weijie~J Su. 2024{\natexlab{b}}.
\newblock A statistical framework of watermarks for large language models: Pivot, detection efficiency and optimal rules.
\newblock \emph{arXiv preprint arXiv:2404.01245}.

\bibitem[{Magister et~al.(2023)Magister, Mallinson, Adamek, Malmi, and Severyn}]{magister-etal-2023-teaching}
Lucie~Charlotte Magister, Jonathan Mallinson, Jakub Adamek, Eric Malmi, and Aliaksei Severyn. 2023.
\newblock \href {https://doi.org/10.18653/v1/2023.acl-short.151} {Teaching small language models to reason}.
\newblock In \emph{Proceedings of the 61st Annual Meeting of the Association for Computational Linguistics (Volume 2: Short Papers)}, pages 1773--1781, Toronto, Canada. Association for Computational Linguistics.

\bibitem[{Mihaylov et~al.(2018)Mihaylov, Clark, Khot, and Sabharwal}]{mihaylov-etal-2018-suit}
Todor Mihaylov, Peter Clark, Tushar Khot, and Ashish Sabharwal. 2018.
\newblock \href {https://doi.org/10.18653/v1/D18-1260} {Can a suit of armor conduct electricity? a new dataset for open book question answering}.
\newblock In \emph{Proceedings of the 2018 Conference on Empirical Methods in Natural Language Processing}, pages 2381--2391, Brussels, Belgium. Association for Computational Linguistics.

\bibitem[{Radford et~al.(2019)Radford, Wu, Child, Luan, Amodei, and Sutskever}]{radford2019language}
Alec Radford, Jeff Wu, Rewon Child, David Luan, Dario Amodei, and Ilya Sutskever. 2019.
\newblock Language models are unsupervised multitask learners.

\bibitem[{See et~al.(2017)See, Liu, and Manning}]{see-etal-2017-get}
Abigail See, Peter~J. Liu, and Christopher~D. Manning. 2017.
\newblock \href {https://doi.org/10.18653/v1/P17-1099} {Get to the point: Summarization with pointer-generator networks}.
\newblock In \emph{Proceedings of the 55th Annual Meeting of the Association for Computational Linguistics (Volume 1: Long Papers)}, pages 1073--1083, Vancouver, Canada. Association for Computational Linguistics.

\bibitem[{Shaib et~al.(2024{\natexlab{a}})Shaib, Barrow, Sun, Siu, Wallace, and Nenkova}]{shaib2024standardizing}
Chantal Shaib, Joe Barrow, Jiuding Sun, Alexa~F Siu, Byron~C Wallace, and Ani Nenkova. 2024{\natexlab{a}}.
\newblock Standardizing the measurement of text diversity: A tool and a comparative analysis of scores.
\newblock \emph{arXiv preprint arXiv:2403.00553}.

\bibitem[{Shaib et~al.(2024{\natexlab{b}})Shaib, Elazar, Li, and Wallace}]{shaib-etal-2024-detection}
Chantal Shaib, Yanai Elazar, Junyi~Jessy Li, and Byron~C Wallace. 2024{\natexlab{b}}.
\newblock \href {https://doi.org/10.18653/v1/2024.emnlp-main.368} {Detection and measurement of syntactic templates in generated text}.
\newblock In \emph{Proceedings of the 2024 Conference on Empirical Methods in Natural Language Processing}, pages 6416--6431, Miami, Florida, USA. Association for Computational Linguistics.

\bibitem[{Shridhar et~al.(2023)Shridhar, Stolfo, and Sachan}]{shridhar-etal-2023-distilling}
Kumar Shridhar, Alessandro Stolfo, and Mrinmaya Sachan. 2023.
\newblock \href {https://doi.org/10.18653/v1/2023.findings-acl.441} {Distilling reasoning capabilities into smaller language models}.
\newblock In \emph{Findings of the Association for Computational Linguistics: ACL 2023}, pages 7059--7073, Toronto, Canada. Association for Computational Linguistics.

\bibitem[{Tafjord et~al.(2018)Tafjord, Clark, Gardner, tau Yih, and Sabharwal}]{Tafjord2018QuaRelAD}
Oyvind Tafjord, Peter Clark, Matt Gardner, Wen tau Yih, and Ashish Sabharwal. 2018.
\newblock \href {https://api.semanticscholar.org/CorpusID:53748665} {Quarel: A dataset and models for answering questions about qualitative relationships}.
\newblock In \emph{AAAI Conference on Artificial Intelligence}.

\bibitem[{Talmor et~al.(2019)Talmor, Herzig, Lourie, and Berant}]{talmor-etal-2019-commonsenseqa}
Alon Talmor, Jonathan Herzig, Nicholas Lourie, and Jonathan Berant. 2019.
\newblock \href {https://doi.org/10.18653/v1/N19-1421} {{C}ommonsense{QA}: A question answering challenge targeting commonsense knowledge}.
\newblock In \emph{Proceedings of the 2019 Conference of the North {A}merican Chapter of the Association for Computational Linguistics: Human Language Technologies, Volume 1 (Long and Short Papers)}, pages 4149--4158, Minneapolis, Minnesota. Association for Computational Linguistics.

\bibitem[{Taori et~al.(2023)Taori, Gulrajani, Zhang, Dubois, Li, Guestrin, Liang, and Hashimoto}]{alpaca}
Rohan Taori, Ishaan Gulrajani, Tianyi Zhang, Yann Dubois, Xuechen Li, Carlos Guestrin, Percy Liang, and Tatsunori~B. Hashimoto. 2023.
\newblock Stanford alpaca: An instruction-following llama model.
\newblock \url{https://github.com/tatsu-lab/stanford_alpaca}.

\bibitem[{Wadhwa et~al.(2024{\natexlab{a}})Wadhwa, Amir, and Wallace}]{wadhwa-etal-2024-investigating}
Somin Wadhwa, Silvio Amir, and Byron~C Wallace. 2024{\natexlab{a}}.
\newblock \href {https://doi.org/10.18653/v1/2024.emnlp-main.349} {Investigating mysteries of {C}o{T}-augmented distillation}.
\newblock In \emph{Proceedings of the 2024 Conference on Empirical Methods in Natural Language Processing}, pages 6071--6086, Miami, Florida, USA. Association for Computational Linguistics.

\bibitem[{Wadhwa et~al.(2024{\natexlab{b}})Wadhwa, Hassanzadeh, Bhattacharjya, Barker, and Ni}]{10.1007/978-3-031-77844-5_13}
Somin Wadhwa, Oktie Hassanzadeh, Debarun Bhattacharjya, Ken Barker, and Jian Ni. 2024{\natexlab{b}}.
\newblock \href {https://doi.org/10.1007/978-3-031-77844-5_13} {Distilling event sequence knowledge from large language models}.
\newblock In \emph{The Semantic Web – ISWC 2024: 23rd International Semantic Web Conference, Baltimore, MD, USA, November 11–15, 2024, Proceedings, Part I}, page 237–255, Berlin, Heidelberg. Springer-Verlag.

\bibitem[{Wei et~al.(2024)Wei, Wang, Schuurmans, Bosma, Ichter, Xia, Chi, Le, and Zhou}]{10.5555/3600270.3602070}
Jason Wei, Xuezhi Wang, Dale Schuurmans, Maarten Bosma, Brian Ichter, Fei Xia, Ed~H. Chi, Quoc~V. Le, and Denny Zhou. 2024.
\newblock Chain-of-thought prompting elicits reasoning in large language models.
\newblock In \emph{Proceedings of the 36th International Conference on Neural Information Processing Systems}, NIPS '22, Red Hook, NY, USA. Curran Associates Inc.

\bibitem[{Wolf et~al.(2020)Wolf, Debut, Sanh, Chaumond, Delangue, Moi, Cistac, Rault, Louf, Funtowicz, Davison, Shleifer, von Platen, Ma, Jernite, Plu, Xu, Scao, Gugger, Drame, Lhoest, and Rush}]{wolf2020huggingfacestransformersstateoftheartnatural}
Thomas Wolf, Lysandre Debut, Victor Sanh, Julien Chaumond, Clement Delangue, Anthony Moi, Pierric Cistac, Tim Rault, Rémi Louf, Morgan Funtowicz, Joe Davison, Sam Shleifer, Patrick von Platen, Clara Ma, Yacine Jernite, Julien Plu, Canwen Xu, Teven~Le Scao, Sylvain Gugger, Mariama Drame, Quentin Lhoest, and Alexander~M. Rush. 2020.
\newblock \href {https://arxiv.org/abs/1910.03771} {Huggingface's transformers: State-of-the-art natural language processing}.
\newblock \emph{Preprint}, arXiv:1910.03771.

\bibitem[{Xu et~al.(2024)Xu, Li, Tao, Shen, Cheng, Li, Xu, Tao, and Zhou}]{xu2024survey}
Xiaohan Xu, Ming Li, Chongyang Tao, Tao Shen, Reynold Cheng, Jinyang Li, Can Xu, Dacheng Tao, and Tianyi Zhou. 2024.
\newblock A survey on knowledge distillation of large language models.
\newblock \emph{arXiv preprint arXiv:2402.13116}.

\bibitem[{Zhang et~al.(2020)Zhang, Kishore, Wu, Weinberger, and Artzi}]{zhang2020bertscoreevaluatingtextgeneration}
Tianyi Zhang, Varsha Kishore, Felix Wu, Kilian~Q. Weinberger, and Yoav Artzi. 2020.
\newblock \href {https://arxiv.org/abs/1904.09675} {Bertscore: Evaluating text generation with bert}.
\newblock \emph{Preprint}, arXiv:1904.09675.

\end{thebibliography}

\clearpage
\appendix
\section*{Appendix}
\label{sec:appendix}

\section{Implementation Details}
We perform all experiments on two NVIDIA A100 GPUs. 
We use publicly available implementations of all models via the {\tt Huggingface} library \cite{wolf2020huggingfacestransformersstateoftheartnatural}. 
For all tasks, we use a learning rate of $3e^{-5}$ and a maximum input length of $1024$. 
We evaluated checkpoints every $500$ steps with early stopping. 
We used a batch size of $12$ for question answering, $2$ for summarization, and $4$ for instruction following. 
Default values were used for all other hyperparameters.

\section{Datasets}
\label{appx:data}
\paragraph{CNN-DailyMail} \cite{see-etal-2017-get} is a large-scale dataset for abstractive text summarization. The dataset consists of online news articles from CNN and the Daily Mail, paired with human-written summaries in the form of bullet points. Each article is accompanied by a summary that captures its key points concisely. The dataset contains over 300,000 news articles and summaries, making it a widely used benchmark for training and evaluating summarization models. The dataset is divided into training, validation, and test sets, with 287,227, 13,368, and 11,490 examples, respectively. 

\paragraph{SumPubMed} \cite{bharti2020} is a large-scale dataset for abstractive biomedical text summarization, derived from PubMed, a leading repository of biomedical literature. Each document in the dataset consists of a full-text research article paired with its structured abstract, enabling the development and evaluation of automatic summarization models in the biomedical domain. The dataset contains over 30,000 articles spanning a diverse range of medical and life sciences topics. 

\paragraph{Rotten Tomatoes} \cite{leone2020rotten} is a dataset of meta-reviews which which synthesize multiple input reviews and and aggregate critic perception of the film. The dataset contains information for 9,095 movies with meta-reviews constructed from 244,000 individual reviews. 

\paragraph{CommonsenseQA} \cite{talmor-etal-2019-commonsenseqa} is a multiple-choice question-answering dataset requiring commonsense knowledge. Each question has five answer choices, with only one correct. The dataset contains 12,102 questions, split into training (9,741), development (1,221), and test (1,140) sets.

\paragraph{OpenbookQA} \cite{mihaylov-etal-2018-suit} is a multiple-choice question-answering dataset that tests reasoning over elementary science facts. Each question includes four answer choices and requires applying scientific facts beyond direct recall. The dataset consists of 5,957 questions, with 4,957 for training, 500 for development, and 500 for testing.

\paragraph{QuaRel} \cite{Tafjord2018QuaRelAD} is a multiple-choice dataset focused on qualitative reasoning, requiring understanding of physical relationships such as speed, force, and heat. Each question has two answer choices and is annotated with a logical representation of the reasoning process. The dataset contains 2,737 questions, split into train (1,911), development (278), and test (548) sets.

\paragraph{Alpaca} \cite{alpaca}  is an instruction-following dataset designed to fine-tune large language models for improved task generalization. Originally containing 52,000 synthetically generated instruction-response pairs from OpenAI's text-davinci-003, this version is downsampled to 10,000 instances, excluding any that contain programming-related code for efficiency. The dataset spans diverse tasks such as reasoning, summarization, and open-ended instruction following.

\begin{figure}
\centering
   \includegraphics[scale=0.379]{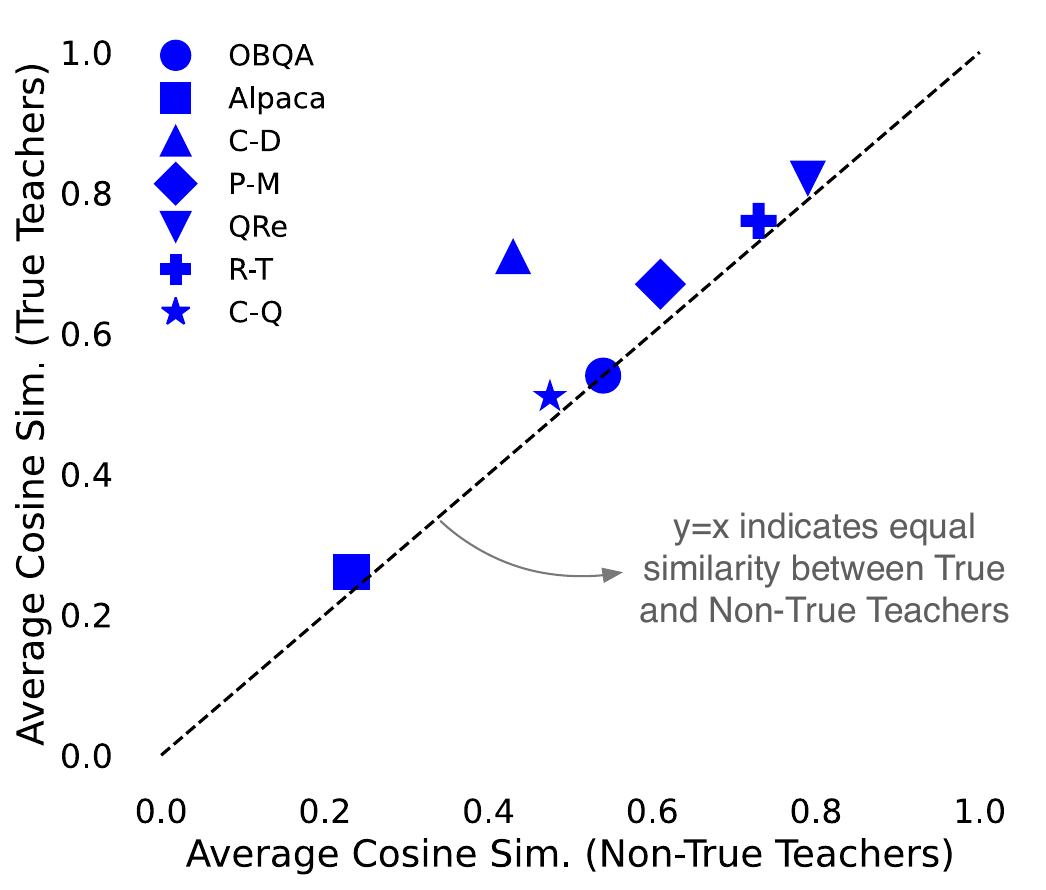}
  \caption{Average cosine similarity of student outputs (across all student models) over Bag-of-Words features with their true teachers vs. other teachers; these features provide little in the way of signal about teachers. }
  \label{fig:sim_fig}
\end{figure}

\section{Additional Results}
\label{appx:add_res} 
\begin{table*}[t]
\centering
\addtolength{\tabcolsep}{7pt} 
\begin{tabular}{@{}llcccc@{}}
\toprule
\multirow{2}{*}{Features}                                                    & Support (\# test instances) \textrightarrow  & 50   & 200  & 1000 & 2000 \\
                                                                             & Data \textdownarrow & \multicolumn{4}{c}{}      \\ \midrule
\multirow{7}{*}{\begin{tabular}[c]{@{}l@{}}BoW \end{tabular}}                & CNN-DailyMail      & 0.44 & 0.44 & 0.45 & 0.45 \\
                                                                             & SumPubMed      & 0.54 & 0.55 & 0.53 & 0.53 \\
                                                                             & Rotten Tomatoes      & 0.38 & 0.37 & 0.37 & 0.37 \\
                                                                             & CommonsenseQA     & 0.43 & 0.40 & 0.41 & 0.41 \\
                                                                             & OpenbookQA     & 0.32 & 0.35 & 0.36 & 0.38 \\
                                                                             & QuaRel      & 0.24 & 0.28 & 0.29 & 0.28 \\
                                                                             & Alpaca   & 0.45 & 0.43 & 0.46 & 0.46 \\ \hline
\multirow{7}{*}{\begin{tabular}[c]{@{}l@{}}N-grams\\(n\textless{}=4)\end{tabular}} 
                                                                             & CNN-DailyMail     & 0.51 & 0.58 & 0.58 & 0.59 \\
                                                                             & SumPubMed      & 0.61 & 0.68 & 0.68 & 0.69 \\
                                                                             & Rotten Tomatoes     & 0.40 & 0.51 & 0.44 & 0.45 \\
                                                                             & CommonsenseQA     & 0.48 & 0.55 & 0.56 & 0.56 \\
                                                                             & OpenbookQA     & 0.45 & 0.48 & 0.48 & 0.49 \\
                                                                             & QuaRel      & 0.43 & 0.47 & 0.50 & 0.50 \\
                                                                             & Alpaca   & 0.52 & 0.55 & 0.56 & 0.56 \\ 
                                                                             \hline
\multirow{7}{*}{\begin{tabular}[c]{@{}l@{}}PoS \\ Templates\end{tabular}}    & CNN-DailyMail      & 0.59 & 0.60 & 0.60 & 0.62 \\
                                                                             & SumPubMed     & 0.68 & 0.70 & 0.71 & 0.72 \\
                                                                             & Rotten Tomatoes      & 0.51 & 0.53 & 0.54 & 0.56 \\
                                                                             & CommonsenseQA     & 0.67 & 0.69 & 0.69 & 0.69 \\
                                                                             & OpenbookQA     & 0.50 & 0.51 & 0.51 & 0.51 \\
                                                                             & QuaRel      & 0.57 & 0.57 & 0.59 & 0.59 \\
                                                                             & Alpaca   & 0.48 & 0.51 & 0.55 & 0.56 \\ 
                                                                             \bottomrule
\end{tabular}
\caption{Classification accuracy of logistic regression models using bag-of-words (BoW), n-grams (n\=1-4), and PoS templates as features across varying support levels (50, 200, 1000, and 2000 test instances). PoS templates consistently outperform BoW and n-grams across most datasets, with performance improving as support increases. The Alpaca dataset presents a marginal exception, where n-grams slightly outperform PoS templates at higher support levels. These results highlight the robustness of PoS templates for distinguishing between teacher models, particularly in high-support settings. Baseline/random accuracy is 0.20..}
\label{tab:add_res}
\end{table*}

To further assess the impact of support levels on classification performance, we extend our evaluation by varying the number of test instances per dataset. Table \ref{tab:add_res} presents results for logistic regression models trained using bag-of-words (BoW), n-grams (up to n=4), and PoS templates as feature representations. We consider support levels of 50, 200, 1000, and 2000 test instances to examine the stability and effectiveness of these representations across different data availability conditions.

Across all datasets and support levels, PoS templates consistently outperform BoW and n-grams in most cases. Notably, the advantage of PoS templates becomes more pronounced as support increases. For instance, in the SumPubMed dataset, accuracy rises from 0.68 at 50 instances to 0.72 at 2000 instances, surpassing both n-grams and BoW at every level. Similarly, in the CommonsenseQA dataset, PoS templates achieve a peak accuracy of 0.69, significantly outperforming BoW (0.41) and n-grams (0.56).

The Alpaca dataset remains an exception, where n-grams achieve slightly better performance (0.56 at 2000 instances) compared to PoS templates (0.56) but exhibit higher variability at lower support levels. This suggests that while PoS templates provide strong structural signals, certain datasets may benefit from richer lexical representations captured by n-grams.

Overall, these findings reaffirm the robustness of PoS templates as a reliable classification feature. Their advantage is particularly evident as more data becomes available, reinforcing the hypothesis that syntactic patterns are distinctive signatures of teacher models. This extended analysis further supports the main section's conclusion that PoS templates offer a scalable and effective alternative to traditional word-based representations.

\end{document}